\documentclass[journal]{IEEEtran}
\usepackage{tabularx,booktabs}
\newcolumntype{C}{>{\centering\arraybackslash}X} 
\ifCLASSINFOpdf
\else
\fi

\usepackage{subfig}
\usepackage[utf8]{inputenc}
\usepackage{todonotes}
\usepackage{mathrsfs}
\usepackage{color,soul}
\usepackage{array,multirow,graphicx}
\usepackage{float}
\usepackage[final]{changes}
 
\begin{document}

\title{CNN-based Methods for Object Recognition with High-Resolution Tactile Sensors}

\author{Juan~M.~Gandarias,~\IEEEmembership{Student Member,~IEEE}, Alfonso~J.~García-Cerezo,~\IEEEmembership{Member,~IEEE} and Jes\'us~M.~Gómez-de-Gabriel,~\IEEEmembership{Member,~IEEE}
\thanks{Submitted on MONTH DAY, 2018. Revised on MONTH DAY, YEAR. An earlier version of this paper was presented at the IEEE SENSORS 2017 Conference and was published in its Proceedings. Digital Object Identifier 10.1109/ICSENS.2017.8234203. This work was supported by the the Spanish project DPI2015-65186-R and the European Commission under grant agreement BES-2016-078237.} 
\thanks{J.M. Gandarias, A.J. Garc\'ia-Cerezo and J.M. G\'omez-de-Gabriel are with the Robotics and Mechatronics Group, Systems Engineering and Automation Department, University of M\'alaga. Escuela de Ingenier\'ias Industriales, M\'alaga, Spain. E-mail: {\tt\small jmgandarias@uma.es}}
\thanks{This paper has supplementary material which includes the tactile dataset and the Matlab code. The material is available in the \replaced{GitHub repository at https://github.com/TaISLab/CNN-based-Methods-for-Tactile-Object-Recognition}}}

\markboth{This is a preprint. The original paper has been published in IEEE SENSORS JOURNAL. DOI: 10.1109/JSEN.2019.2912968}%
{Gandarias \MakeLowercase{\textit{et al.}}: Tactile Object Recognition with Convolutional Neural Networks}

\maketitle

\begin{abstract}
Novel high-resolution pressure-sensor arrays allow treating pressure readings as standard images. Computer vision algorithms and methods such as Convolutional Neural Networks (CNN) can be used to identify contact objects. In this paper, a high-resolution tactile sensor has been attached to a robotic end-effector to identify contacted objects. Two CNN-based approaches have been employed to classify pressure images. These methods include a transfer learning approach using a pre-trained CNN on an RGB-images dataset and a custom-made CNN (TactNet) trained from scratch with tactile information. The transfer learning approach can be carried out by retraining the classification layers of the network or replacing these layers with an SVM. Overall, 11 configurations based on these methods have been tested: 8 transfer learning-based, and 3 TactNet-based. Moreover, a study of the performance of the methods and a comparative discussion with the current state-of-the-art on tactile object recognition is presented.

\end{abstract}


%
\IEEEpeerreviewmaketitle


 


%
%
\section{Introduction}

\IEEEPARstart{S}{ense} of touch is essential for human beings to perform complex tasks such as object recognition. 
As for humans, tactile information is also useful for robotics systems \cite{LuoShanBimboJoaoDahiyaRavinder2017RoboticReview,Trujillo-Leon2018TactileWheelchairs}. 
Nowadays research studies in robotics are focusing on making robots more similar to humans. That includes providing the robot tactile perception \cite{Bartolozzi2016RobotsTouch,Jamone2015HighlyHand}. However, tactile sensing is not enough, and some level of intelligence is required.
This way, tactile perception is still a fundamental problem in field robotics that has not be solved so far \cite{Roncone2016PeripersonalSkin}. To overcome this, two points have to be addressed: first, obtaining sensing tactile information, and second, having the cognitive capabilities to process that information \cite{Dahiya2010TactileHumanoids}. 

\begin{figure}
\centering
\includegraphics[width = 1 \columnwidth]{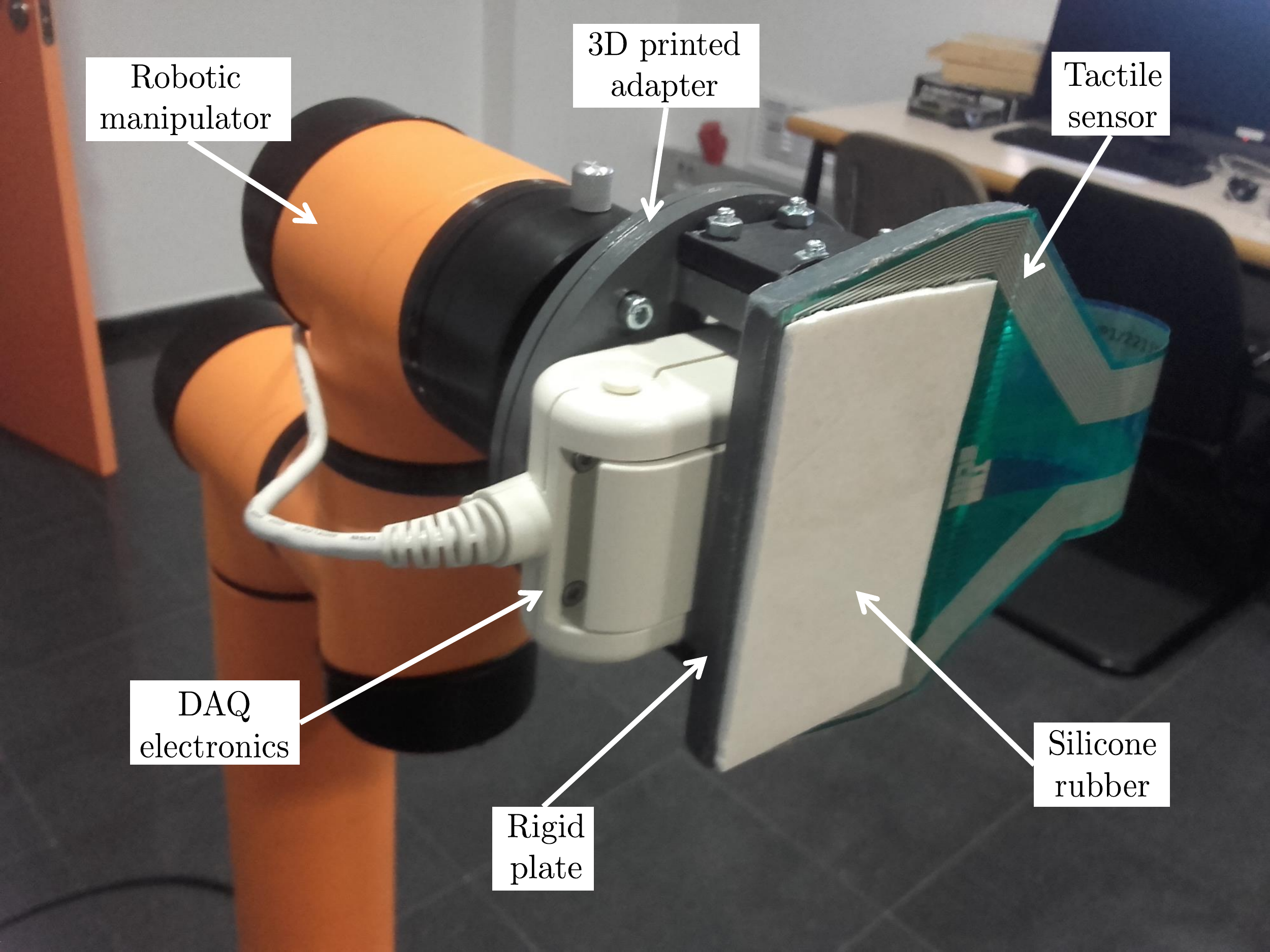}
\caption{The tactile end-effector attached on the robotic manipulator \textit{AUBO OUR-i5} as a platform to touch objects and collect data.}
\label{fig:system}
\end{figure}

On the one hand, multiple tactile sensors \added{(as the one presented in Fig.~\ref{fig:system})} have been employed in robotics manipulators and grippers \cite{Gandarias2018EnhancingInteraction, Chitta2011TactileManipulation} to carry out a large variety of applications such as slippage detection \cite{James2018SlipSensor, Romeo2017SlippageSensors}, tactile object recognition \cite{Gandarias2017HumanSensor, Luo2016IterativeRecognition} and surface classification \cite{Yuan2017DesignClassification, Hoelscher2015EvaluationRecognition, li2013sensing}. Tactile sensors estimate the pressure by measuring different physical magnitudes. Thus, the information given depends on the nature of the transducer. For example, typical piezoresistive tactile sensors produce a pressure image where each pixel represents the pressure measured by each tactel \cite{vidal2011three}, while hall-effect sensors are used in combination with magnets and elastic materials  
\cite{Chathuranga2016MagneticSensor} to estimate the pressure in larger areas. Other technologies have also been used to estimate contact pressure from other physical magnitudes such as light \cite{Ward-Cherrier2018TheMorphologies}, air pressure \cite{Gong2017ARobots} or capacitance \cite{Maiolino2013ARobots}.

On the other hand, novel Artificial Intelligence (AI) techniques can be used for interpreting the information perceived by tactile sensors. 
Machine learning methods such as Gaussian Processes \cite{YiZhengkun}, Bayesian approaches \cite{Corradi2015BayesianSensor}, k-mean clustering and Support Vector Machines (SVM) \cite{Albini2017HumanInteractions} or k-Nearest Neighbour (kNN) \cite{Luo2015NovelRecognition} have been used for distinguishing contacts.
Novel Convolutional Neural Networks (CNNs) are also acquiring excellent results in multiple applications such as visual object recognition \cite{Krizhevsky2012ImageNetNetworks}. These methods can be used for recognizing objects contacted through tactile sensors \cite{Cao2018End-to-EndEnsemble,Shibata2017,gandarias2017tactile}.

\begin{table}
\caption{\added{Main differences between RGB and tactile images. *The number of features depends on each image itself, but in general terms it could be affirmed that tactile imprints have lower number of features than RGB images.}}
\centering 
\begin{tabularx}{\columnwidth}{@{}l*{3}{C}c@{}}
\toprule
\textbf{Factors}	& \textbf{RGB} & \textbf{Tactile}\\
\midrule
\midrule
Spatial resolution & $1.6 \times 10^5$  & $ 1.4 \times 10^3 $ \\
\midrule
Color & Yes & No \\
\midrule
Background & Yes & No \\
\midrule
Magnitude  & Light intensity & Pressure \\
\midrule
Number of features*  & High & Low \\
\midrule
Object in image & Whole object & Depends on the size \\
\midrule
Area of detection & Large & Small \\
\midrule
Depth of field & Large & Small \\
\midrule
Type of perception & Passive & Active \\
\midrule
 & Shading, texture, & \\
3D visual signs & focus, movement,  & No \\
 & perspective & \\
\midrule
Amount of data & $1.3 \times 10^6$ & $1.1 \times 10^3$ \\

\bottomrule
\label{tab:differences}
\end{tabularx}
\end{table}


It is also possible to transfer the knowledge from other domains to reduce the efforts of data collection, building a CNN model, and training the new network from scratch (i.e. transfer learning). This way, some applications can take advantage of the developments made for similar fields (e.g. pressure versus visual image classification),  using previous calibration data reduce training time for new devices, and labelling form one classifier can be reused for different input sets \cite{pan2010survey}. 
CNN for video image recognition automatically detect features and provide image invariances such as rotation, scale, and brightness \cite{Gandarias2018EnhancingInteraction}.
In particular, transfer learning is specially useful when the amount of training data from the domain of interest is limited. In \cite{mihalkova2009transfer}, the transfer when the amount of target data is minimal is studied. 

\begin{figure}
\centering
\includegraphics[width = 1\columnwidth]{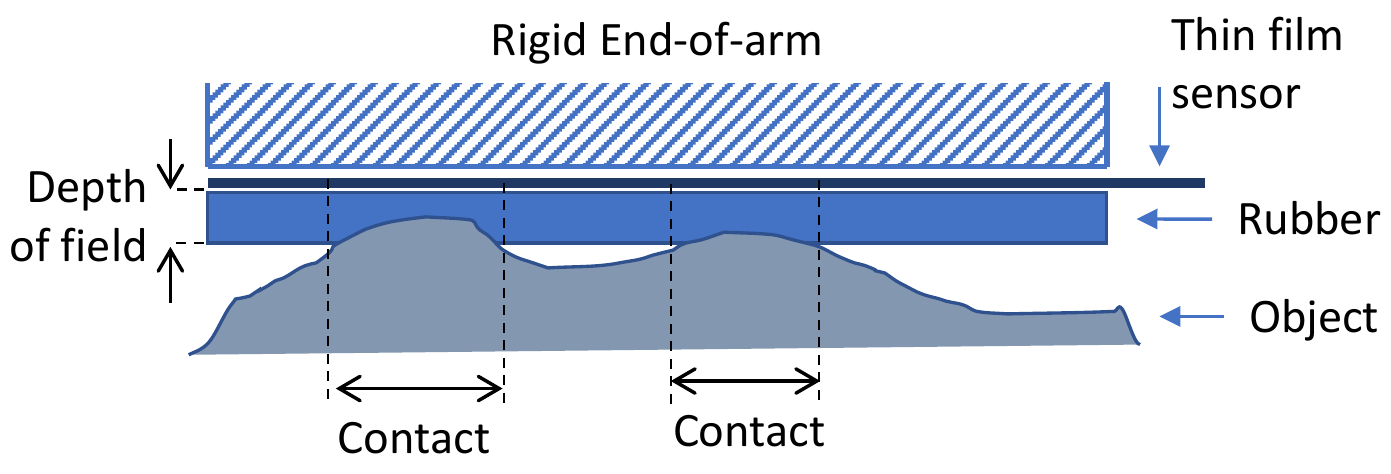}\\
a)\\
\vspace{0.15cm}
\includegraphics[width = 0.48\columnwidth]{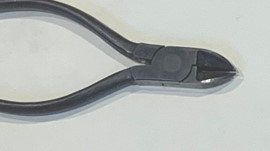}
\includegraphics[width = 0.48 \columnwidth]{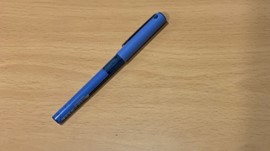}\\ 
\vspace{0.15cm}
\includegraphics[width = 0.48 \columnwidth]{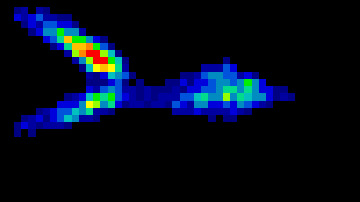}
\includegraphics[width = 0.48 \columnwidth]{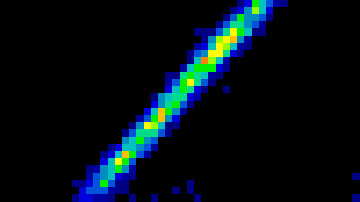}\\ 
\vspace{0.15cm}
\includegraphics[width = 0.48 \columnwidth]{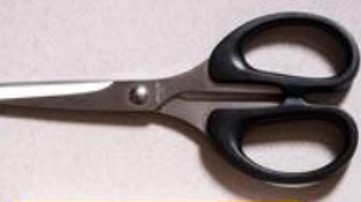}
\includegraphics[width = 0.48 \columnwidth]{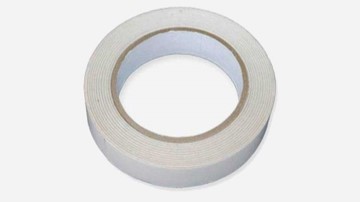}\\ 
\vspace{0.15cm}
\includegraphics[width = 0.48 \columnwidth]{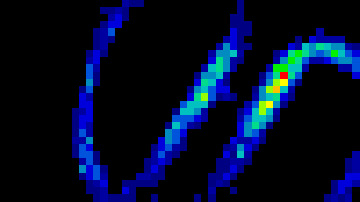}
\includegraphics[width = 0.48 \columnwidth]{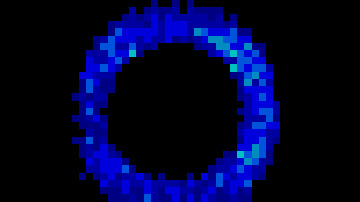}\\
b)
\caption{\added{Differences between visual and pressure images a) Illustration of the depth-of-field in pressure images. b) Examples of video (top) vs pressure (bottom) images for common objects (from left to right and top to bottom: pliers, pen, scissors and tape).}} 
\label{fig:pressure_map}
\end{figure}

The available deep CNNs trained for image recognition can have millions of weights, and the size of the layers can be bigger than the actual input data from the new application domain. The application of these traditional CNNs to tactile object recognition may be not efficient due to the shortage of features in tactile images in comparison with RGB images, and also to the lower resolution of tactile images \added{among other factors. The main differences between RGB and tactile images can be seen in Fig.~\ref{fig:pressure_map} and are summarized in Table~\ref{tab:differences}}. Moreover, using these networks for embedded applications, with computational and energy constraints, is not suitable.
For embedded applications in domains with a smaller input data size (i.e. lower resolution images) or a lower number of output labels, a custom CNN built from scratch may be done off-line in a different computer) with an eventually limited amount of training data has to be addressed using data augmentation.

\added{This paper is focused on the application of tactile perception techniques for practical autonomous robots, that have limited computational power and energy. In that sense, this work provides comparative results of \replaced{eleven}{ a huge variety of} CNN-based approaches to recognize pressure images in terms of recognition rate and computational load.}
Two main approaches are presented: transfer learning from CNNs pre-trained in a dataset of RGB images, and training a custom CNN from scratch using tactile information only. 
Tactile data is collected by a high-resolution piezoresistive tactile sensor which has been attached to a robotic manipulator as shown in Fig. \ref{fig:system}. \added{CNN-based techniques, commonly used in visual object recognition tasks, are used to classify tactile data. In particular, different CNN-based methods are evaluated in a 22-classes experiment. Transfer learning approach is compared with a custom-made CNN in terms of accuracy and classification time. The customized CNN has been named \textit{TactNet} and includes 3 configurations: TactNet-4, TactNet-6 and TactResNet.}
The results are discussed, and the most beneficial situation for using one method or another is determined based on the experimentation, considering the recognition rate and the classification time.  
Besides, these approaches are analyzed and compared against the current state-of-the-art.
Therefore, this paper can also be used as a guide for further readers whose intend to use CNNs to classify pressure images. Furthermore, the data and the methods have been made available in the \replaced{GitHub}{\textit{Code Ocean}} cloud-based platform, so they can be used with different methods to compare new approaches while providing reproducibility to this work.

%
%
\section{Related Work}
\label{sec:related}

In literature, two main approaches can be followed: surface/material identification and object recognition.
Material identification refers to the extraction of information for the superficial properties of an object like roughness, texture, stiffness or thermal conductivity \cite{li2013sensing, Feng2018ActiveObjects, Kaboli2018RobustSkin, Baishya2016RobustLearning}. Sequential data is important to identify the properties of objects. The interpretation of tactile information as time-series of data is presented in multiple works about material identification \cite{Madry2014ST-HMP:Data, Liu2016ObjectMethods}. An exploratory motion is carried out with a robotic arm in \cite{kerzel} to obtain dynamic information with a 2D force sensor (normal and shear forces). The control of the actuator is critical to maintain the pressure against the contact material constant and obtain trustworthy information. In that case, a multi-channel neural network is used and a high accuracy is achieved.

On the other hand, object recognition concerns to distinguish objects by its shape \cite{jamali2011majority, Liu2012ASensor, Martinez-Hernandez2017FeelingHand}. Apart from the approach, most of these works are based on the use of AI methods to identify tactile information. The use of one method or another depends on the type of the sensed information. 

Related works used Neural Networks and Deep Learning techniques to recognize objects \cite{Khasnobish2012Object-shapeNetwork, Schmitz2014TactileDropout}. This way, the benefits of using CNNs can be exploited. The translation-invariant property allows the recognition independently of the position of the object in the image \cite{Lawrence1997FaceApproach}. In \cite{Albini2017HumanInteractions}, a CNN-based method is developed to recognize in-contact human hand with artificial skin. The method takes advantage of the translation-invariance of the networks, being able to identify both right and the left hands, indistinctly. The use of Deep Learning with dropout to reduce overfitting is presented in \cite{Schmitz2014TactileDropout}. This work also describes the benefits of including both kinesthetic and tactile information to object shape recognition and raises the differences between using planar or curved tactile sensors. 
Luo et al. also propose a novel algorithm which synthesizes both kinesthetic and tactile information, forming a 4D point cloud of the object with the label numbers of the tactile features as an additional dimension to the 3D sensor positions \cite{Luo2016IterativeRecognition}.  

Other idea consists of applying multi-modal techniques \cite{Falco2017Cross-modalExploration}. Fusing haptic and visual data generally presents better results than using a single kind of information \cite{gao2016deep}. In \cite{Zheng2016DeepInformation} a multi-modal deep learning method based on a CNN is presented. The network takes the information sensed with an accelerometer and the corresponding image of the surface material as inputs and estimates the contact material.

The information acquired by a pressure sensor array can be represented as a matrix, the same way as an image. Multiple studies treat tactile data as images \cite{ Gandarias2018EnhancingInteraction, Albini2017HumanInteractions, Luo2015NovelRecognition, luo2015tactile, schneider2009object}. Most of them are based on the same two steps: feature extraction from pressure images and obtaining a classifier based on those features \cite{Luo2014RotationSensor}. 
An existing solution uses a variant of the \textit{Scale Invariant Feature Transform} (SIFT) descriptor as a feature extractor and a supervised \textit{k-Nearest Neighbour} (kNN) algorithm to get the classifier \cite{Luo2015NovelRecognition}. However, multiple touches are needed to obtain good results in the classification task.

Pressure data extracted from tactile arrays can also be considered as a series of images. A flexible high-resolution tactile sensor is used in \cite{Shibata2017} to classify food textures. A CNN is used with a sequence of pressure images, taken when an experimental device crushes the food imitating the movement of a mouth closing and biting food. Although time series of tactile data is important to record the behaviour of gel-like food, authors found that using only two pressure images produces similar results.

%
%
\section{Methodologies}
\label{sec:method}

\subsection{Tactile images}
%

\begin{figure}
	\centering
    \includegraphics[width = 0.3\columnwidth]{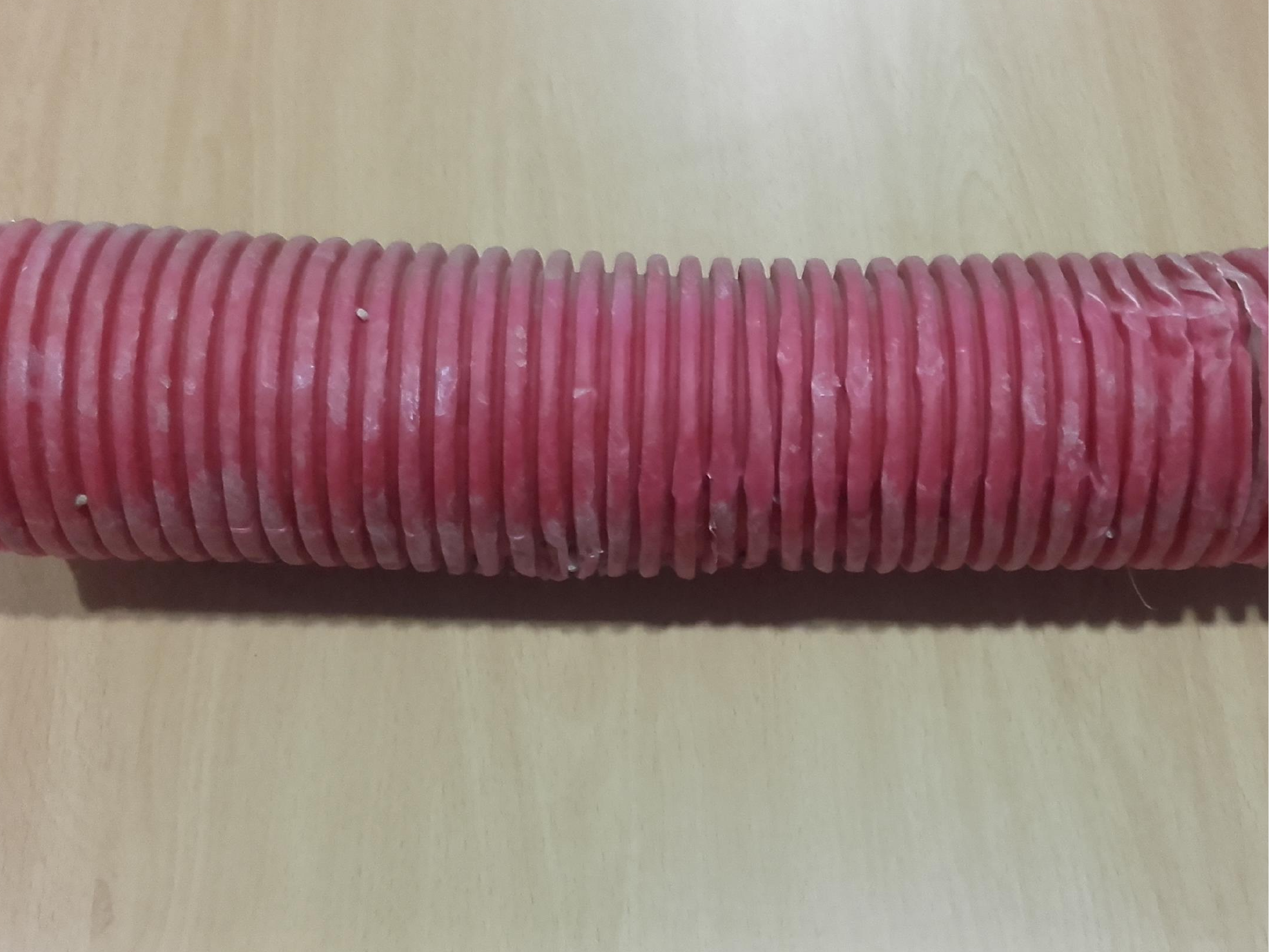}\\ 
    a) Cable pipe\\
    \vspace{2mm}
    \includegraphics[width = 0.3\columnwidth]{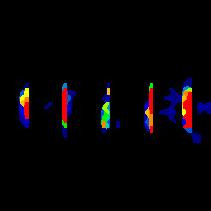} 
    \includegraphics[width = 0.3\columnwidth]{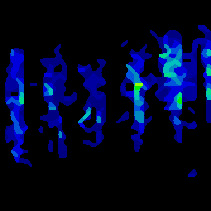} 
    \includegraphics[width = 0.3\columnwidth]{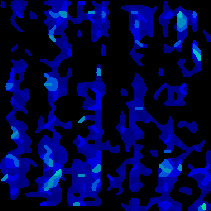} \\
    b) Rigid \hspace{1cm} c) Semi-rigid \hspace{1cm} d) Flexible\\
    \includegraphics[width = 1.0\columnwidth]{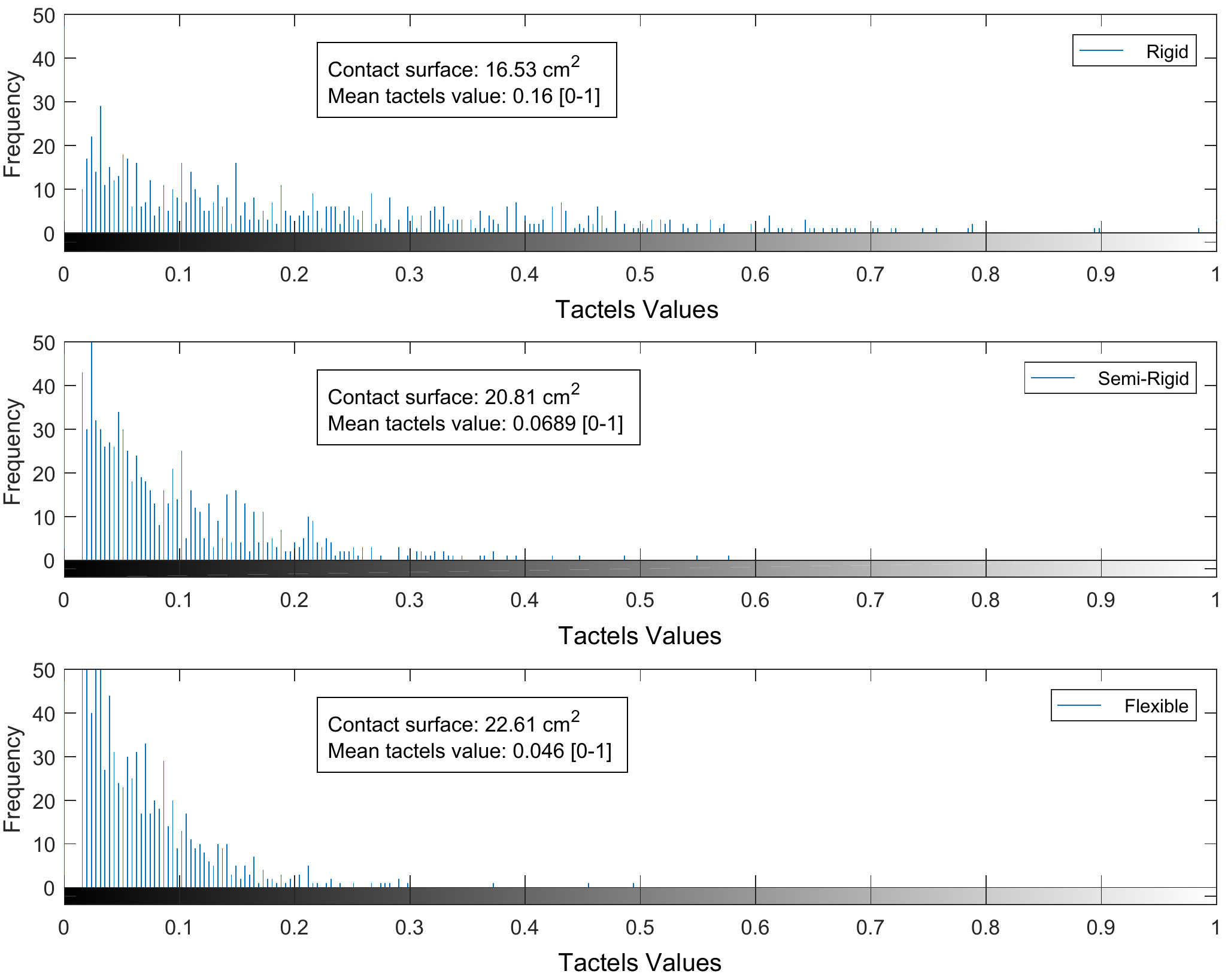} \\
    e) Histograms of the pressure images\\
    \caption{\added{Pressure images have limited field-of-depth and depend on the compliance of the object and the sensor. a) Sensed object. b) Pressure image from a rigid sensor. c) Pressure image from a semi-rigid sensor. c) Pressure image from a flexible sensor. e) Histograms of the pressure images, including contact surface and mean tactel value for images b) c) and d).}}
    \label{fig:histogram}
\end{figure}

\added{Pressure images have limited field-of-depth and depend on the compliance of the object and the sensor. In Figure \ref{fig:histogram}, the pressure images obtained with three different sensor compliances, from the same plastic pipe are shown. Also, quantitative measurements have been done and included in the histograms in Fig.\ref{fig:histogram}(e), that shows how pressure images obtained from flexible sensors that bend around the objects have higher contact area, but lower dynamic range than the images obtained with a flat rigid sensor \cite{gandarias2017tactile}.
The rigid configuration has higher pressure values than the others and avoid bending the thin-film sensor. This arrangement (as seen in Figure \ref{fig:system}) includes a layer of silicon rubber to further protect the film and enhance the depth of field (see Fig. \ref{fig:pressure_map}) of the sensor, and has been chosen for the rest of the experiments in this paper.}

\subsection{Transfer Learning}
Transfer learning consists of using a trained neural network for a different purpose for which it was created. Thus, a CNN trained in a large amount of RGB images can be used to classify pressure images. This idea takes advantage of the particular roles played by each parts of a CNN. The first convolutional layers of the network learn to extract features from images, whereas the last layers learn to classify the input data. Hence, this approach consists of using the convolutional layers of a pre-trained CNN and replacing the classification layers with a custom classifier. 

\begin{figure}
\centering
\includegraphics[width = 1 \columnwidth]{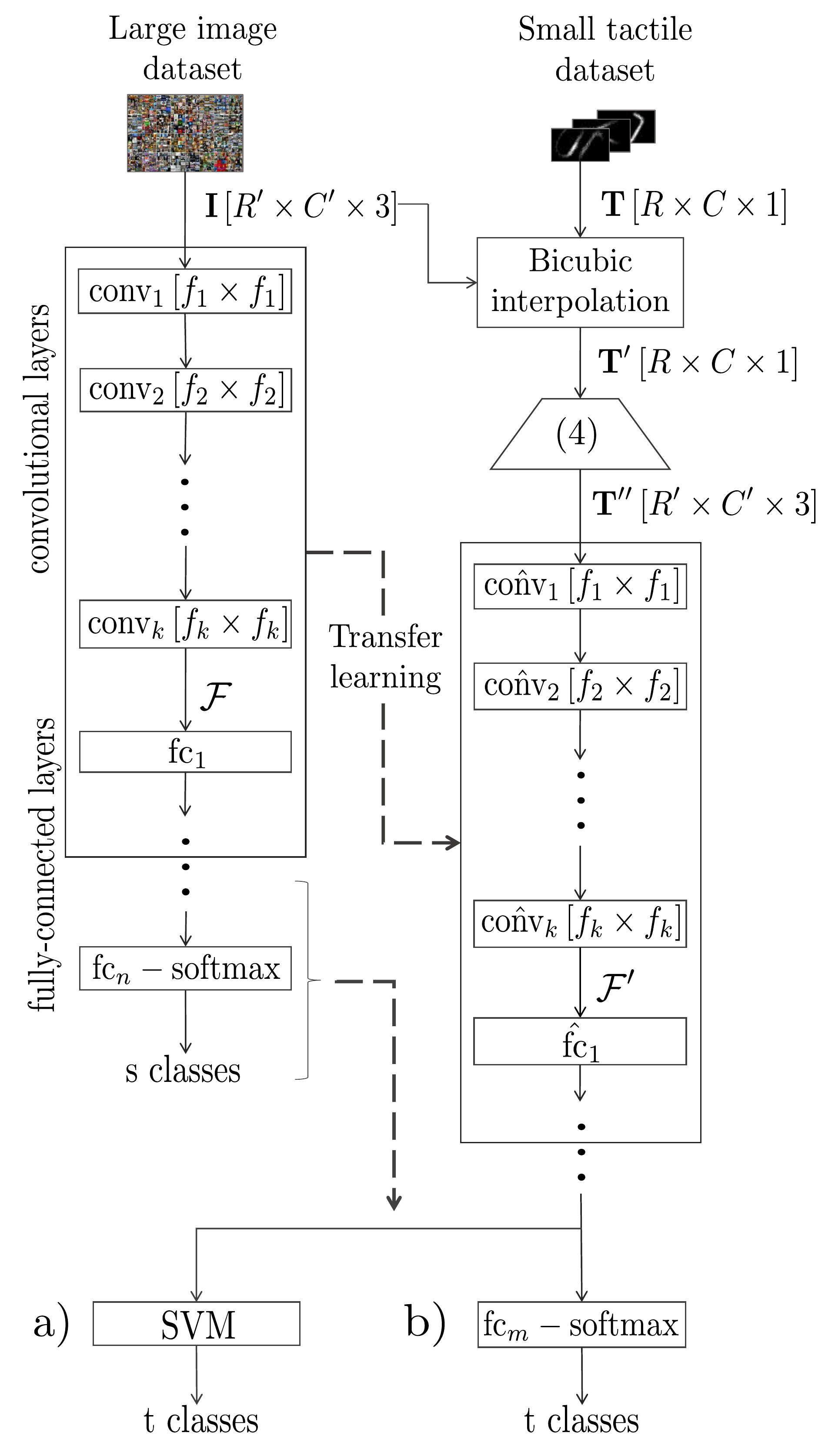}
\caption{Structure of the transfer learning approach. The replacement of the classification layers can be carried out using an SVM a) or fine-tuning the classification layers of the network b).} 
\label{fig:transfer_learning}
\end{figure}

The structure of this approach is presented in Fig.~\ref{fig:transfer_learning}. 
The first step consists of training a CNN in a large image dataset. This CNN has two distinguishable parts: the firsts convolutional layers $\left( \mathscr{C} = \left[\textrm{conv}_1, \ldots, \textrm{conv}_k \right] \right) $ with filters sizes $\left[f_1 \times f_1 \right] , \ldots, \left[f_k \times f_k \right]$; and the lasts fully-connected layers $\left( \mathscr{F} = \left[\textrm{fc}_1, \ldots, \textrm{fc}_n \right] \right) $. After the training process, the convolutional part of the network learns to extract features $\mathcal{F}$ from RGB images, while the fully-connected layers learn to classify this features. 

Considering this factor, the first part of the trained network $\left( \hat{\mathscr{C}} = \left[\hat{\textrm{conv}}_1, \ldots, \hat{\textrm{conv}}_k \right] \right)$ which has learn to extract features from common RGB images $\mathcal{F}$, can be exploited to extract features from pressure images $\mathcal{F}'$. In this step, some trained fully-connected layers $\left( \hat{\mathscr{F}} = \left[\hat{\textrm{fc}}_1, \ldots, \hat{\textrm{fc}}_t \right] \textrm{with} \, t<n \right) $ can also be employed.

However, as the structure of the network has been set previously, the size of the input tactile data $\left(\mathbf{T}\left(r,c\right) \, \forall \, r=1,...,R, \, c=1,...,C \right)$ has to match the size of the RGB data used for training the convolutional layers $\left(\mathbf{I}\left(r',c',z\right) \, \forall \, r'=1,...,R', \, c'=1,...,C', \, z=1,2,3 \right)$. 

To use the transfer learning approach and take advantage of the pre-learned parameters, the whole structure of the network must be maintained. Hence if the pre-trained network has been trained to classify $\mathbf{I} \left[ R' \times \ C' \times 3 \right]$ images, this image size has to be kept. Thus it is needed to resize tactile images $\mathbf{T} \left[R \times C \times 1 \right]$ to $ \mathbf{T''}\left[ R' \times \ C' \times 3 \right]$.

Then the bicubic interpolation method has been applied. This method obtains better quality results than others, although it needs larger amount of calculation \cite{Hisham2017AnMethod}. However, the time of this calculation can be disregarded in comparison with the neural network computation time. Each pixel $\mathbf{T'}\left(r',c'\right)$ of the resized image is obtained from its 16-neighbour pixels applying the equation (\ref{eq:bicubic}):

\begin{equation}
\mathbf{T'} \left(r',c'\right) = \sum_{i=0}^{3}\sum_{j=0}^{3} \left[r_i \times c_j\right] \, \mathbf{T}\left(i,j\right),
\label{eq:bicubic}
\end{equation}

Where $\mathbf{T}\left(i,j\right)$ is the 16-neighbour matrix, and $r_i$ and $c_j$ can be obtained by the equations (\ref{eq:ri}) and (\ref{eq:cj}):

\begin{equation}
r_i = \prod_{k=0, k \neq i}^3 \frac{r'-\left[S_r\times\left(x+k\right)\right]}{\left[S_r\times\left(x+i\right)\right] - \left[S_r\times\left(x+k\right) \right]}, 
\label{eq:ri}
\end{equation}

\begin{equation}
c_j = \prod_{k=0, k \neq i}^3 \frac{c'-\left[S_c\times\left(y+k\right)\right]}{\left[S_c\times\left(y+i\right)\right] - \left[S_c\times\left(y+k\right) \right]} ,
\label{eq:cj}
\end{equation}

$x$ and $y$ are the values of each row and column divided by the scale factor $S_r = R/R'$ and $S_c = C/C'$ respectively. 
Finally, according to equation \ref{eq:extrapol}, the matrix $\mathbf{T}' \left[R' \times C' \times 1 \right]$
is used in each of the 3 channels of $\mathbf{T}''$ to match the size of the original RGB image $\mathbf{I}\left[ R' \times C' \times 3 \right]$, resulting a 3-channel tactile image $\mathbf{T}''\left[ R' \times C' \times 3 \right]$.

\begin{equation}
\mathbf{T}''\left(r',c',z\right) = \mathbf{T}'\left(r',c'\right), z \in \left[1,2,3\right] ,
\label{eq:extrapol}
\end{equation}

Then, as shown at the bottom of the scheme in Fig.~\ref{fig:transfer_learning}, the classification layers of the network can be replaced by two classifiers. On one hand, the substitution of the classification layer by an SVM is contemplated in Fig.~\ref{fig:transfer_learning} a). On the other hand, a simple idea is to train the classification layers again (fine-tuning) on a tactile dataset Fig.~\ref{fig:transfer_learning} b).

\subsection{TactNet}
Other idea consists of creating a CNN from scratch and training it from a pressure image dataset, however, a large amount of tactile data is needed. For this work, a CNN for tactile information classification (TactNet) has been created. In this study, this network is configured with 4 and 6 plain layers, following the architecture of the AlexNet \cite{Krizhevsky2012ImageNetNetworks}, and with 6 layers including residual convolutions, following the structure of the ResNet \cite{he2016deep}. 

These networks are composed by a set of convolutional $\left( \mathscr{C} = \left[\textrm{conv}_1, \ldots, \textrm{conv}_k \right] \right)$ or residual convolutions $\left( \mathscr{R} = \left[\textrm{res}_1, \ldots, \textrm{res}_k \right] \right)$ which learn to extract tactile features $\mathcal{F}_t$ from pressure images. A batch normalization layer with $\epsilon = 10^{-4}$ is introduced after each convolution, followed by a rectified linear unit (ReLU) to introduce non-linearities. Some layers are alternated with a max-pooling layer with a stride of $2$. Finally, a set of $n$ fully connected layers $\left( \mathscr{F} = \left[\textrm{fc}_n \right] \right)$ learn to classify the input data.

The details of the three different configurations of TactNet that have been considered in this work are explained below. The networks were implemented in \textit{Matlab R2018b} using the \textit{Neural Network Toolbox}. The implementation details can be found at the \replaced{GitHub }{Code Ocean} repository 
\footnote{https://github.com/TaISLab/CNN-based-Methods-for-Tactile-Object-Recognition}

\subsubsection{TactNet-4}
\begin{figure}
\centering
\includegraphics[width = 1 \columnwidth]{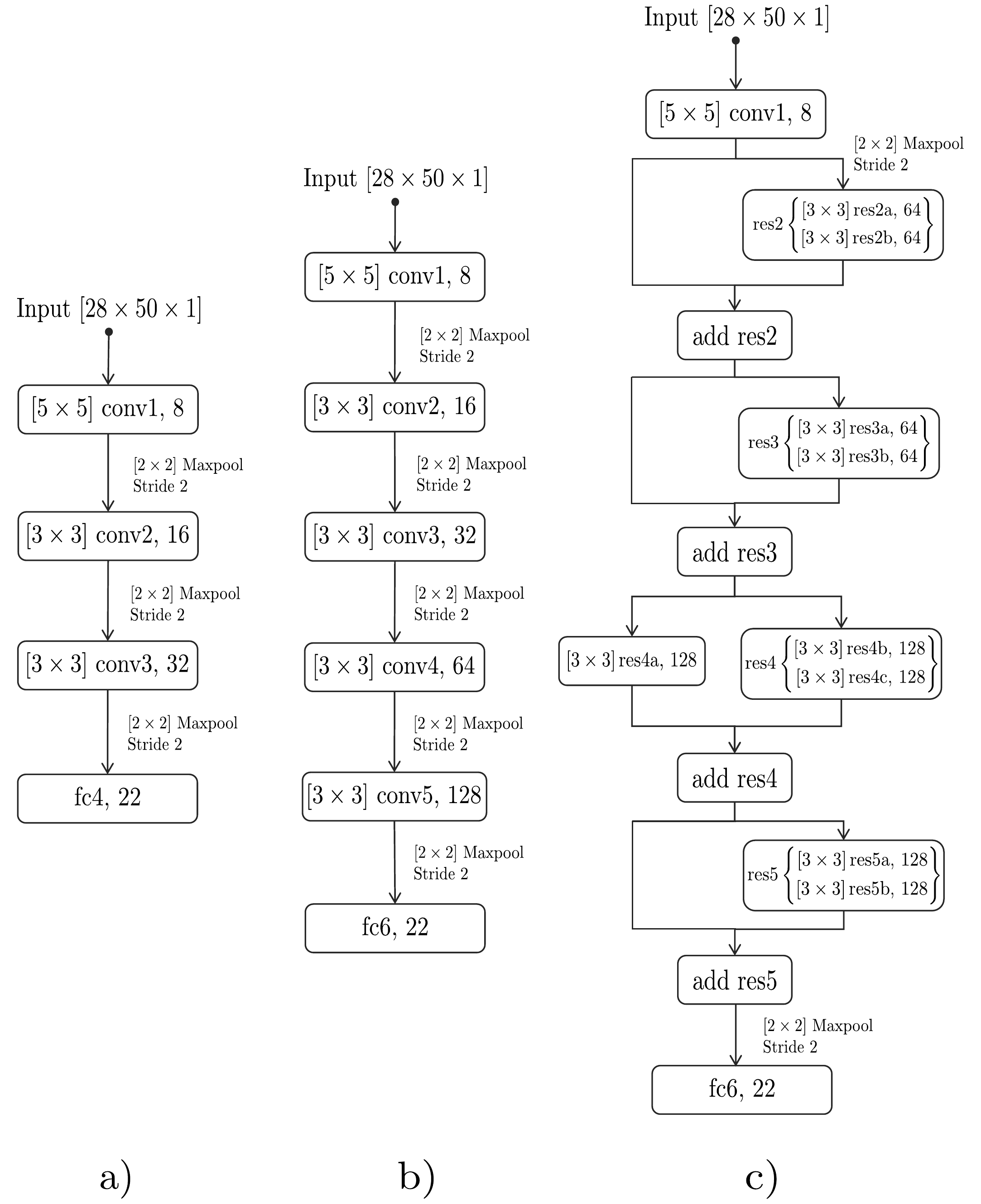}
\caption{Structure of the TactNet configurations: TactNet-4 a), TactNet-6 b) and TactResNet c)} 
\label{fig:TactNet}
\end{figure}

This is the simplest configuration of TactNet. The network is composed by 3 convolutional layers $\left( \mathscr{C} = \left[\textrm{conv}_1, \textrm{conv}_2, \textrm{conv}_3 \right] \right)$ with filters sizes $\left[5 \times 5 \right], 8$ \ , $\left[3 \times 3 \right], 16$  and $\left[3 \times 3 \right], 32 $ respectively; and a fully-connected layer $\left( \mathscr{F} = \left[\textrm{fc}_4 \right] \right)$ with 22 neurons followed by a softmax-layer to classify the input tactile data and gives the likelihood of belonging to each class. The structure of this network is shown in Fig.~\ref{fig:TactNet} a).

\subsubsection{TactNet-6}
In this case, the network is formed by 5 convolutional layers $\left( \mathscr{C} = \left[\textrm{conv}_1, \textrm{conv}_2, \textrm{conv}_3, \textrm{conv}_4, \textrm{conv}_5 \right] \right) $ with filters sizes $\left[5 \times 5 \right],8 \ , \left[5 \times 5 \right],16  \ , \left[3 \times 3 \right],32 \ , \left[3 \times 3 \right],64 $ and $\left[3 \times 3 \right], 128$ respectively. After that, 1 fully-connected layer $\left( \mathscr{F} = \left[\textrm{fc}_6 \right] \right)$ with 22 neurons, and a softmax-layer to get the probabilities. The structure of this network is shown in Fig.~\ref{fig:TactNet} b).

\subsubsection{TactResNet}
This network is formed by 1 convolutional layer $\left( \mathscr{C} = \left[\textrm{conv}_1 \right] \right)$, 4 residual layers $\left( \mathscr{R} = \left[\textrm{res}_2, \textrm{res}_3, \textrm{res}_4, \textrm{res}_5 \right] \right) $, and 1 fully connected layer $\left( \mathscr{F} = \left[\textrm{fc}_6 \right] \right)$ followed by a softmax-layer to get the probabilities. The filters and specifications of this network are presented in Fig.~\ref{fig:TactNet} c).

%
%
\section{Experimental Setup}
\label{sec:setup}

\subsection{Description of the experiments}
The performance of the proposed methodologies is evaluated based on the recognition rates achieved in an experiment with objects from 22 classes. Tactile information from these objects has been collected to carry out the learning processes.

Tactile data acquisition is a critical task as training data determines the learning process, therefore the tactile sensor is placed in contact with different objects manually. The operator decides when a sample (pressure image) is valid or not based on the visual information provided by the user interface of the \textit{I-Scan} data acquisition software, provided by \textit{Tekscan}. With this software the operator sees the pressure map in real-time. 

In Fig.~\ref{fig:pressure_map}, an example of tactile imprints of four objects used for the experiment is presented. The real-time pressure map provided by the data acquisition software is formed by 15 values (colors) image. The color of each pixel depends on the pressure applied in each tactel. Hence, the minimum pressure (black) corresponds to $0 \, \textrm{Pa}$, whereas the maximum pressure (red) corresponds to the maximum pressure admitted by the sensor.

The experimental implementation consists of testing and comparing 11 different classification procedures. Following the transfer learning approach, 8 methods have been considered: 4 for the SVM classifier and 4 for re-training the last layers of the network as a classifier \added{(see Fig.~\ref{fig:transfer_learning})}. The pre-trained networks used for extract features are AlexNet \cite{Krizhevsky2012ImageNetNetworks}, ResNet \cite{he2016deep},  SqueezeNet \cite{iandola2016squeezenet} and VGG16 \cite{Simonyan2014VeryRecognition}.  On the other hand, the 3 configurations of TactNet presented in Fig.~\ref{fig:TactNet} have been also contemplated. The 11 methods are summarized in Table~\ref{tab:class_summary}. Moreover, the SURF-SVM method described in \cite{Gandarias2017HumanSensor} has been included in the experiment as a comparison point to show the enhancement of the CNN-based methods against a traditional approach that doesn't use deep learning.

\begin{table}
\caption{\added{Summary of the CNN-based techniques used for the classification experiment (TL = Transfer Learning, conv = Convolutional layer, res = Residual layer, SM = Softmax).}}
\centering 
\begin{tabularx}{\columnwidth}{@{}l*{4}{l}c@{}}
\toprule
\textbf{Approach} & \textbf{Name}	& \textbf{Feat. Extractor} & \textbf{Classifier}\\
\midrule
\parbox[t]{2cm}{\multirow{4}{*}{TL-SVM (Fig.\ref{fig:transfer_learning}a)}}
& SqueezeNet-SVM & SqueezeNet & SVM\\
& AlexNet-SVM & AlexNet & SVM\\
& ResNet-SVM & ResNet & SVM\\
& VGG-SVM & VGG16 & SVM\\
\midrule
\parbox[t]{2cm}{\multirow{4}{*}{TL-NN (Fig.\ref{fig:transfer_learning}b)}}
& SqueezeNet-NN & SqueezeNet & 1conv-SM\\
& AlexNet-NN & AlexNet & 2fc-SM\\
& ResNet-NN & ResNet & 1conv-SM\\
& VGG-NN & VGG16 & 3fc-SM\\
\midrule
\parbox[t]{2cm}{\multirow{3}{*}{TactNet (Fig.\ref{fig:TactNet})}}
& TactNet-4 & 3conv & 1fc-SM\\
& TactNet-6 & 5conv& 1fc-SM \\
& TactResNet & 1conv-4res & 1fc-SM\\
\bottomrule
\label{tab:class_summary}
\end{tabularx}
\end{table}

To carry out the experiment, two different hardware systems are used. A GPU NVidia GeForce GTX 1050 Ti with 4 GB of RAM and a CPU Intel Core i7-7700HQ CPU @ 2.80 GHz. The methods have been trained using both systems and the classification time has been measured on each one.

\subsection{Sensor specifications}
The tactile end-effector employed for the experiments, shown in Fig.~\ref{fig:system}, uses a high-resolution tactile sensor that has been attached to the 6 DOF robotic arm \textit{AUBO Our-i5}. An own-designed 3D printed part is used for coupling the sensor to the robotic manipulator. This component has been manufactured using Fused Deposition Modelling (FDM) 3D printing in PLA (PolyLactic Acid) plastic.

The data acquisition system is constituted by the tactile sensor model 6077, the \textit{Evolution Handle} and the \textit{I-Scan} software, which are provided by \textit{Tekscan} (South Boston, MA, USA). The high-resolution tactile-array has a total of 1400 pressure sensels (also known as tactels or taxels). Each tactel has a size of $53.3 \times 95.3 \textrm{mm}$, conforming a set of resistive pressure sensors with density $27.6 \, \textrm{tactels} / \textrm{cm}^2$ distributed in a matrix of $28$ rows by $50$ columns. The main features of the sensor are detailed in Table~\ref{tab:sensor_specifications}.

\begin{table}
\caption{Specifications of the high-resolution tactile sensor model 6077 from \textit{Tekscan}.}
\centering 
\begin{tabularx}{\columnwidth}{@{}l*{2}{C}c@{}}
\toprule
\textbf{Parameter}	& \textbf{Value}\\
\midrule
Max. pressure & $34$  KPa\\
Number of tactels & $1400$  \\
Tactels density & $27 \, \textrm{tactels}/\textrm{cm}^2$\\
Temperature range & $\textrm{-}40\, ^\circ$C to $ \textrm{+} 60\, ^\circ \textrm{C}$  \\
Matrix height & $53.3 \, \textrm{mm}$  \\
Matrix width & $95.3 \, \textrm{mm}$  \\
Thickness & $ 0.102 \, \textrm{mm}$   \\
\bottomrule
\label{tab:sensor_specifications}
\end{tabularx}
\end{table}

The tactile sensor is covered by a silicone rubber as a contact interface which protects the sensor and conducts external forces. A silicon rubber pad (sourced by \textit{RS} with code $733-6713$) provides soft contact between the sensor and the objects.

\subsection{Dataset}

For training the models, the use of two datasets is considered. On one hand, CNNs used in transfer learning models are pre-trained on more than a million images and can classify images into 1000 object categories. The ImageNet dataset is formed by common RGB images and is used in the ImageNet Large-Scale Visual Recognition Challenge \cite{russakovsky2015imagenet}. 

On the other hand, a dataset formed by tactile images has been collected. This dataset is used to train the TactNet models and the classifiers of the transfer learning methods. A total of 1100 pressure images have been used to feed each method. These images are divided into 22 classes labelled as: \textit{Adhesive, allen key, arm, ball, bottle, box, branch, cable, cable pipe, caliper, can, finger, hand, highlighter pen, key, pen, pliers, rock, rubber, scissors, sticky tape and tube.}

Learning processes require three subsets of tactile data: training set, validation set and test set. The training set is composed by 704 images, 32 images for each label, whilst the validation and test set are composed by 176 and 220 images respectively. 

TactNet models are trained using data augmentation techniques. Exploiting the translation, rotation and scale invariant traits introduced by data augmentation, the models are able to recognize touched objects independently of their location in the image, orientation or contact pressure \cite{kauderer2017quantifying}. Note that ImageNet dataset already includes data augmentation techniques, therefore these invariant qualities are also presented in transfer learning models.

In the case of the learning process of the different configurations of TactNet, the data augmentation techniques considered have been reflections, rotations and translations in X and Y axis. Hence, the amount of tactile data available for training the networks is 4224, while the validation and test set are 1056 and 1320 respectively.

%
%

\section{Results}
\label{sec:results}

In this section, the results achieved by the applied methods according to the experiment statement are shown and explained. 

\begin{figure}
\centering
\includegraphics[width = 1 \columnwidth]{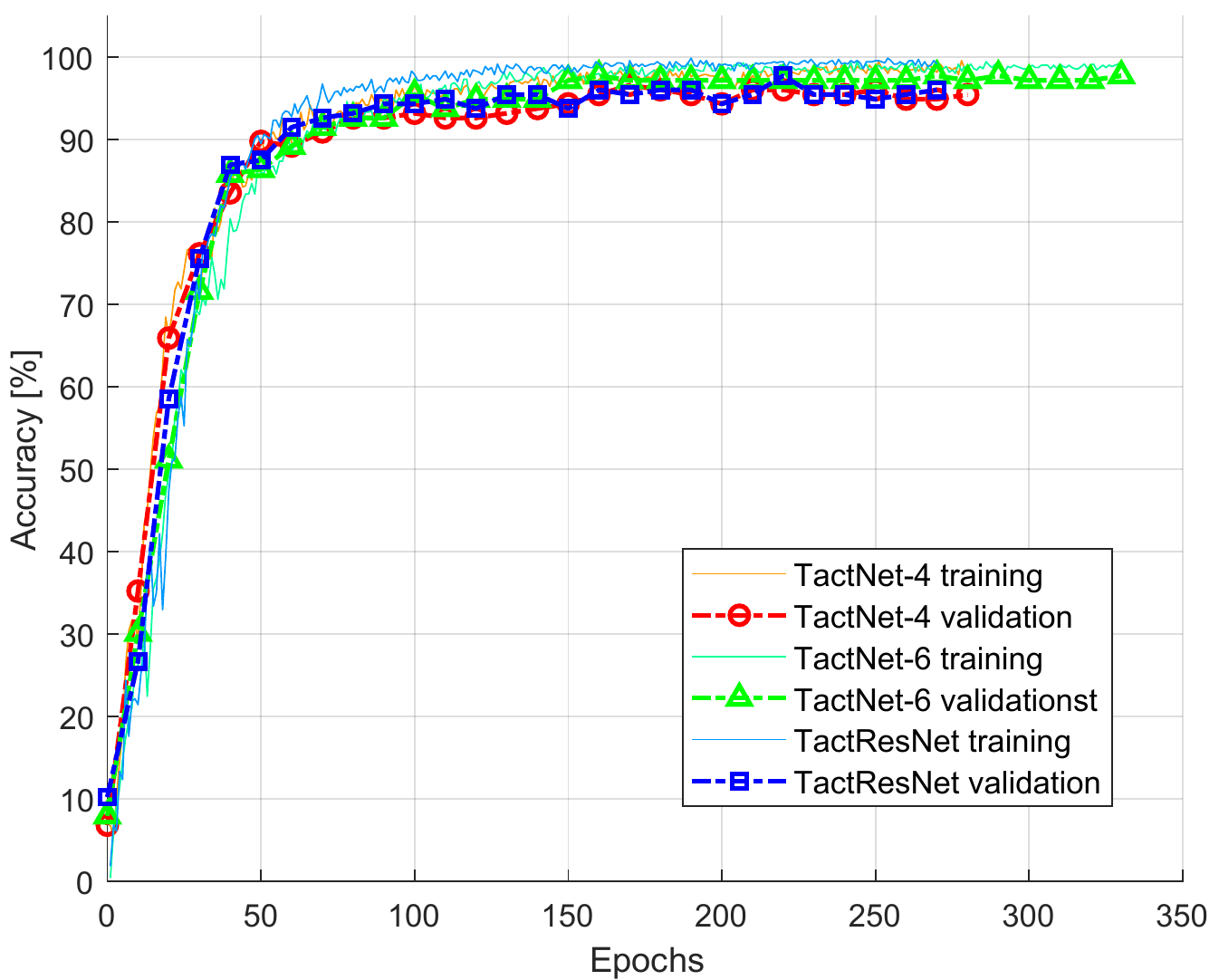}
\caption{Learning process of the 3 TactNet configurations.}
\label{fig:learning_process}
\end{figure}

Fig.~\ref{fig:learning_process} shows the learning processes of each TactNet configuration from scratch. In this graph, it is shown that the accuracy achieved by each configuration is close to $100 \%$.

\begin{figure}
\centering
\includegraphics[width = 1 \columnwidth]{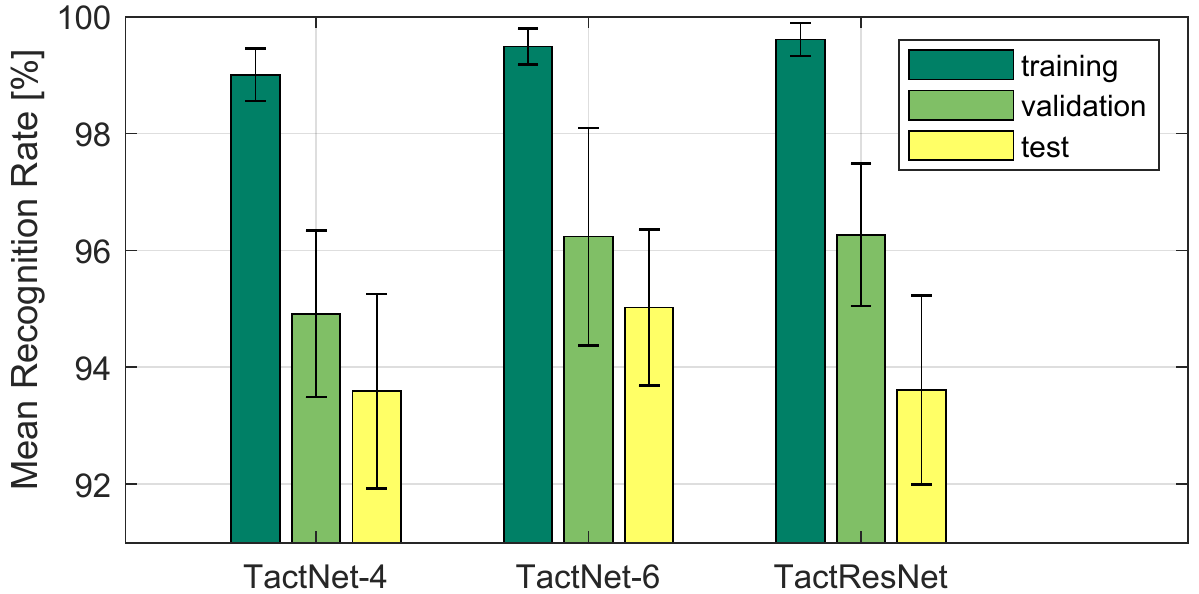}
\caption{Accuracy achieved after the 20 samples of learning process by each TactNet configurations with the training, validation and test sets. The error bars represent the one standard deviation.}
\label{fig:mean_recognition_tactilenet}
\end{figure}

As the images used for the training process affect the performance of the network, \replaced{20 training sets have been randomly collected from the dataset to train each method, then 20 validation and 20 test sets were formed, also randomly, with the rest of the data not used for training.}{each method has been trained 20 times with a random collection of images for the training, validation and test sets.}
The \added{accuracy achieved} by each TactNet configurations over the 20 training samples is presented in Fig.~\ref{fig:mean_recognition_tactilenet}. Each bar represents the mean recognition rate achieved by each configuration with the training, validation and test sets. The error bars represent the one standard deviation over the 20 samples. As expected, the recognition rate achieved with the training rate is the highest while the one achieved with the test set is the lowest. It is also shown that the mean recognition rate is over $93\%$ for all the methods.

\begin{figure}
\centering
\includegraphics[width = 1 \columnwidth]{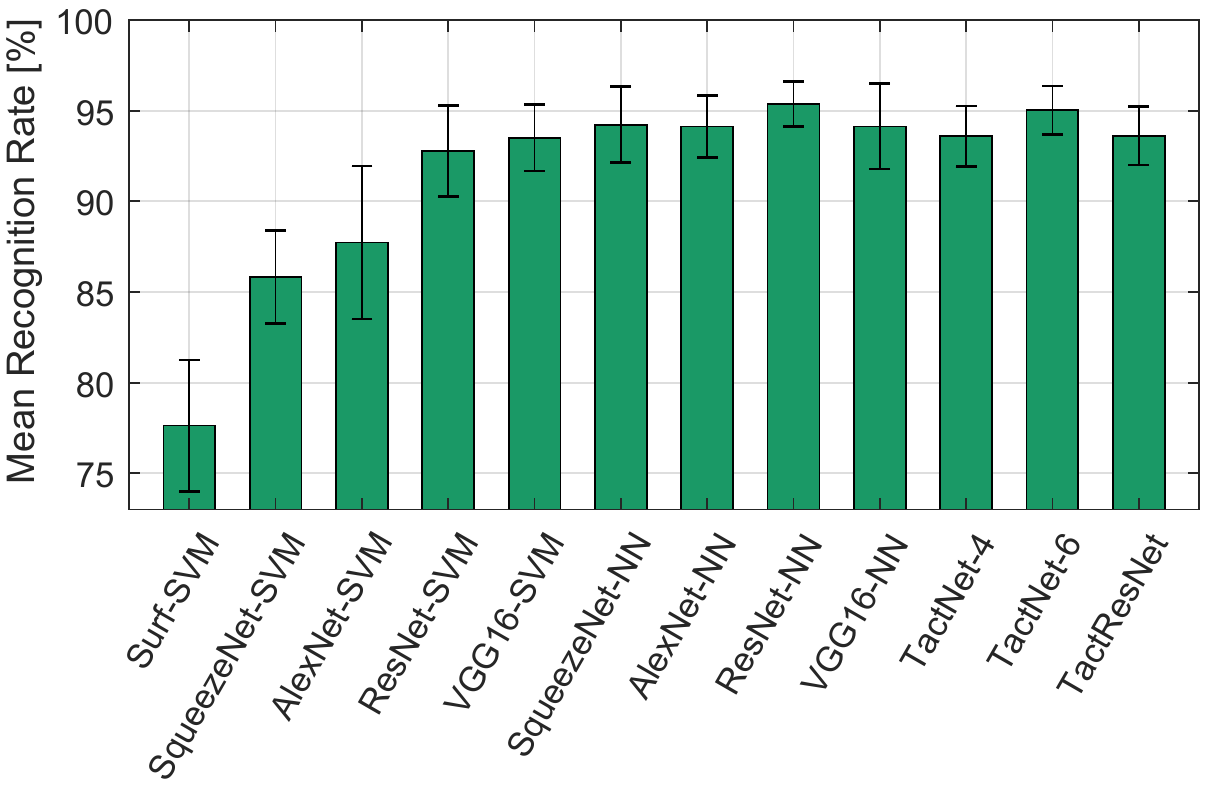}
\caption{Comparison of the recognition rate achieves by each method. The error bars represent the one standard deviation.}
\label{fig:accuracy_comparison}
\end{figure}

Fig.~\ref{fig:accuracy_comparison} presents a comparison graph of the mean recognition rate achieved by the transfer learning approaches and TactNet configurations. The method SURF-SVM has also been evaluated to be compared against the CNN-based approaches. As in the case of the TactNet, the training process of the transfer learning methods and SURF-SVM has been carried out with 20 samples of randomized data.

\begin{figure}
\centering
\includegraphics[width = 1 \columnwidth]{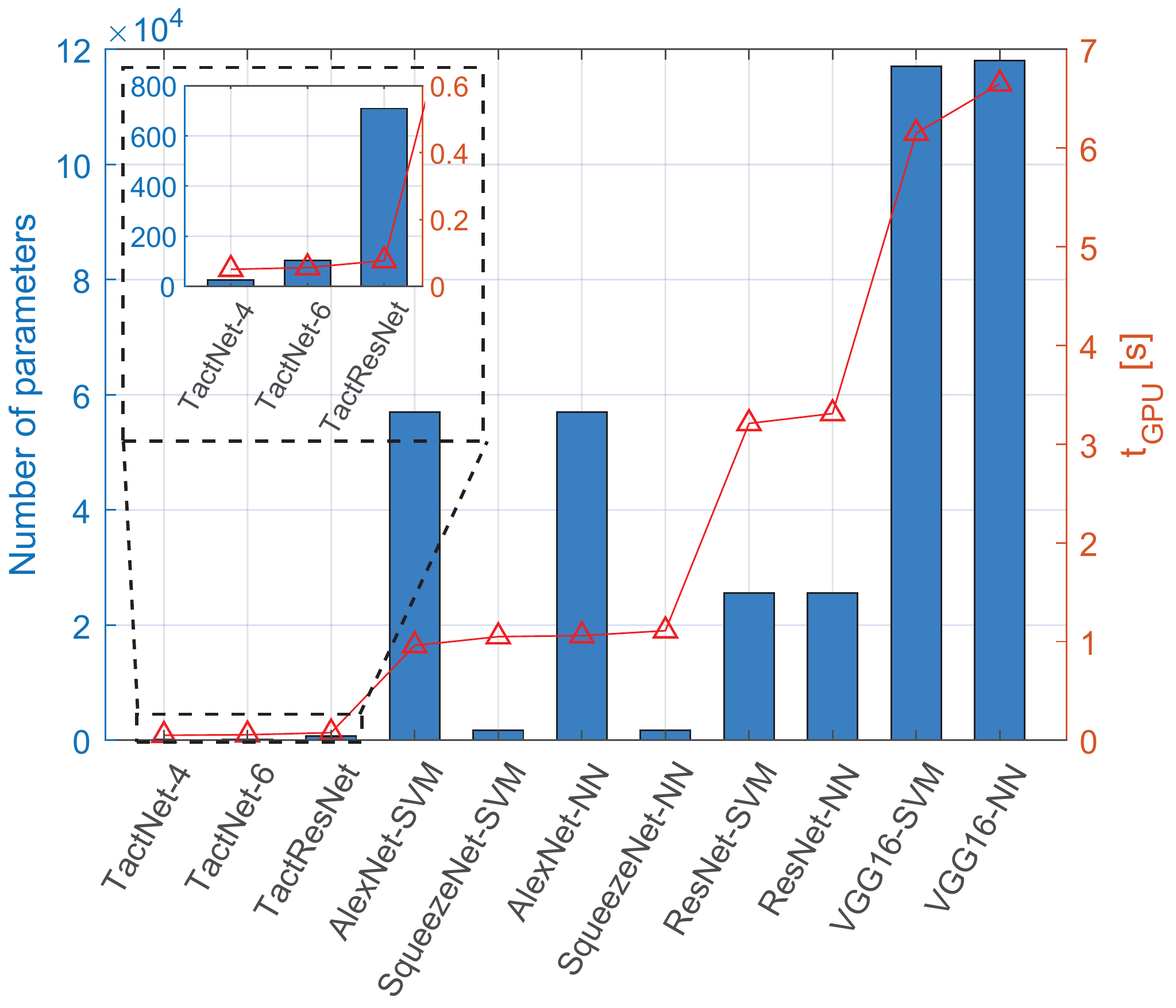}
\caption{Comparison of the number of parameter of each method, and the computation time for each one. The computation time considered in this graph corresponds to the classification time of the methods running on a NVidia GeForce GTX 1050Ti GPU.}
\label{fig:Nparm_time}
\end{figure}

Another aspect that has to be considered is the classification time. This is highly dependent on the number of parameters of a CNN and the hardware, and according to \cite{Canziani2016AnApplications} the number of parameters of a network is proportional to the number of operations, the size and the inference time. Therefore, the higher number of parameters a network has, the more memory and processing time is required.
Fig. \ref{fig:Nparm_time} shows the number of parameters of the CNN used in this work and the computation time for each one running on the NVidia GPU.

\begin{figure}
\centering
\includegraphics[width = 1 \columnwidth]{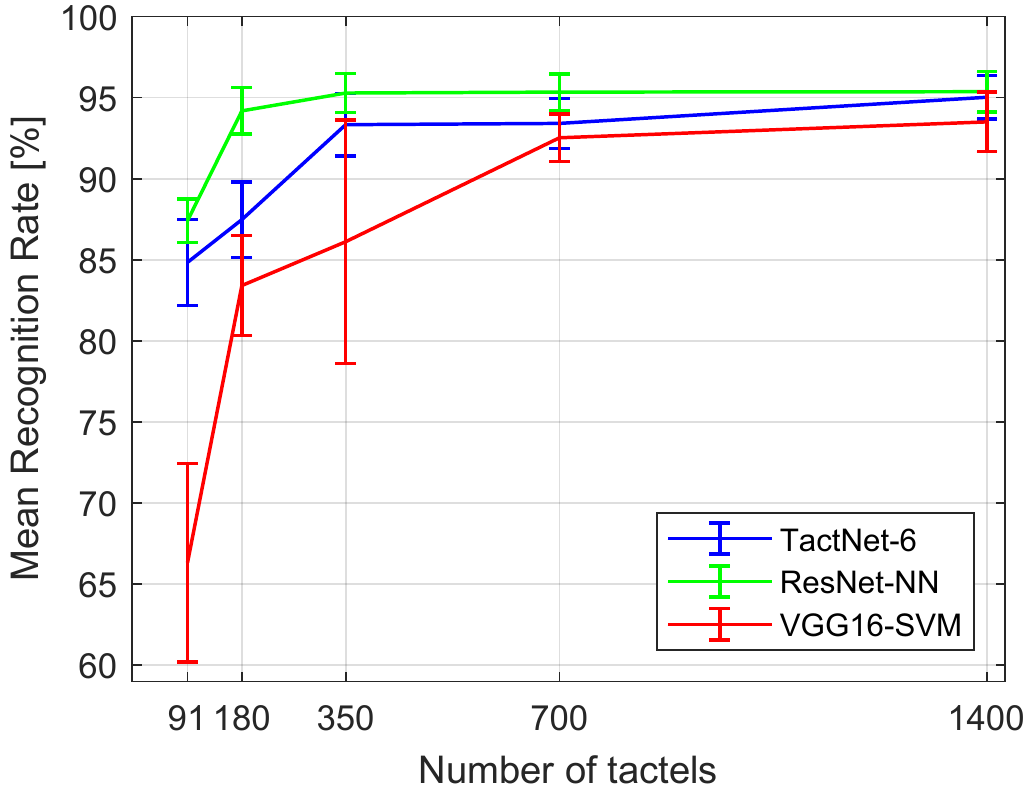}
\caption{\added{Comparison of the performance of the best methods of each CNN-based approach (see Table~\ref{tab:class_summary} and Table~\ref{tab: comparison_CNN}) for different sensor's resolutions using one touch only. The tactels' distribution for each resolution is: $91-[7 \times 13]$, $180-[10 \times 18]$, $350-[14 \times 25]$, $700-[20 \times 35]$ and $1400-[28 \times 50]$.}}
\label{fig:spatial_resolution}
\end{figure}

\added{On the other hand, the spatial resolution of the sensor plays an important role on the performance of the task. Therefore, a classification experiment with the best methods of each approach: VGG16-SVM for transfer learning with SVM (see Fig.\ref{fig:transfer_learning}a), ResNet-NN for transfer learning with fine-tuning (see Fig.\ref{fig:transfer_learning}b) and TactNet6 for TactNet (see Fig. \ref{fig:TactNet}b), has been carried out using approximately $\times 1/16$, $\times 1/8$, $\times 1/4$, $\times 1/2$ and $\times 1$ resolution of the tactile sensor. The results are presented in Fig.~\ref{fig:spatial_resolution}. The experiment is done using only one touch to classify the objects, and the resolution of the sensor is decreased by software using the bicubic interpolation (see section~\ref{sec:method}B). The structure of TactNet-6 is changed to match the dimensions of the input data for each resolution, and some hyperparameters are changed to ensure good training process. This changes are not necessary in case of transfer learning approaches as the input data is resized again to match the original input size of the network. }

%
%
\section{Analysis and Discussion}
\label{sec:discussion}

A summary of the results is presented in Table~\ref{tab: comparison_CNN}. This shows that both transfer learning and TactNet obtain high accuracy rates on tactile object recognition, and that using a CNN-based approach can raise the mean recognition accuracy up to $17.72\%$ against SURF-SVM in the best case (ResNet-NN). In fact, the worst CNN method in terms of accuracy (SqueezeNet-SVM) has an improvement of $8.08\%$ against SURF-SVM.
The recognition rate in the case of transfer learning depends on the CNN and the classifier, however if a large CNN is used and the classifier consists of fine-tuning the fully-connected layers of the network, the mean recognition rate and the classification time increases. Nevertheless, the classification times of the TactNet configurations are significantly shorter than in transfer learning methods.
These results show the better performance of CNN-based method against traditional approaches, especially when using TactNet configurations, which present similar recognition rates to the best transfer learning methods, but need substantially shorter classification times. \added{Moreover, a further reduction in computing time can be obtained for the TactNet approach by using lower resolutions.}

\begin{table}
\caption{\added{Performance comparison between methods used for tactile object recognition}.} 

\centering 
\begin{tabularx}{\columnwidth}{@{}l*{5}{c}c@{}}
\toprule
 \textbf{ Model } & \textbf{Parameters} & \textbf{Accuracy[\%]} & \textbf{$\textrm{t}_{\scriptsize \textrm{GPU}}$ [s]} & \textbf{$\textrm{t}_{\scriptsize \textrm{CPU}}$ [s]}\\
\midrule
SURF-SVM  &  -  & 77.64 &  0.025 & 0.029 \\
\midrule
SqueezeNet-SVM & 1.7M & 85.82 & 1.05 & 7.934\\
AlexNet-SVM &  57 M & 87.73 & 0.96 & 4.141\\
ResNet-SVM & 25.6M & 92.77 & 3.21 & 23.848\\
VGG16-SVM  & 117 M & \textbf{93.50} &  6.15 & 73.355\\
\midrule
SqueezeNet-NN & 1.7M & 94.23 & 1.11 & 8.515\\
AlexNet-NN  &  57 M & 94.14 & 1.06 & 4.982\\
ResNet-NN & 25.6M & \textbf{95.36} & 3.31 & 27.039\\
VGG16-NN   &  118 M & 94.14 & 6.65 & 64.488 \\
\midrule
TactNet-4  & 25 k & 93.59 &  \textbf{0.051} &  \textbf{0.094}\\
TactNet-6  &  104 k & \textbf{95.02} &  \textbf{0.056} & \textbf{0.103}\\
TactResNet & 790 k & 93.61 & \textbf{0.077} & \textbf{0.465}\\
\bottomrule
\label{tab: comparison_CNN}
\end{tabularx}
\end{table}

\added{As different researchers are using different hardware, data and methods, to compare the proposed method with existing related works, a table with the  qualitative/quantitative features (see Table~\ref{tab: state_art}) of relevant methods that use tactile information as images, has been included.
Different methodologies have been employed, mostly applying computer vision techniques for extracting features, followed by a classifier, as well as the use of Artificial Neural Networks (ANN). It can be seen that a large variety of tactile sensors, number of classes, number of touches and type of objects can be used, and all these aspects have an influence on the accuracy independent of the followed approach. In general, the larger number of classes, the smaller accuracy, and the larger number of touches, the higher accuracy}. 

\begin{table*}
 \caption{State-of-the-art on tactile object recognition. Due to the nature of this work, and the variety of the approaches, only those studies which treat tactile data as common images are considered. The last three conform the methods presented in this paper.}
\label{tab: state_art}
\begin{tabularx}{\textwidth}{@{}l*{7}{llcccl}c@{}}
\toprule
Year  & Method & Tactile sensor & Nº of classes & Accuracy [\%] & Nº of touches & Type of objects\\ 
\midrule
2009 & Bag of Features \cite{schneider2009object} & 2 fingers tactile array & 21 & 84.6 & 10 & Household + industrial \\
2012 & PCA + SOM + BoK + ANN \cite{Navarro2012HapticHands} & Schunk dexterous hand & 10 & 78.88 & 4 & Household \\
2014 & Stacked DAEs \cite{Schmitz2014TactileDropout} & Twendy-one hand & 20 & 88.00 & 4 & Household \\
2015 & Tactile-SIFT \cite{Luo2015NovelRecognition} & $[6 \times 14] $ tactile array & 18 & 91.33 & 15 & Household \\
2015 & Cascade of 2 NN \cite{Cretu2015ComputationalRecognition} & $[8 \times 16]$ tactile array & 25 & 96.00 & 1 & Symbols (Letters) \\
2016 & Zernike moment + iClap \cite{Luo2016IterativeRecognition} & $[6 \times 14] $ tactile array & 20 & 85.36 & 20 & Household\\
2016 & kernel sparse coding \cite{Liu2016ObjectMethods} & Barret hand & 10 & $> 90.00$ & 1 & Household \\
2017 & SURF + k-mean + SVM \cite{Gandarias2017HumanSensor} & $[28 \times 50]$ Tactile array & \centering 8 & 80.00 & 1 & Household + body parts\\
2017 & CNN + SVM \cite{Gandarias2017HumanSensor} & $[28 \times 50]$ Tactile array & 8 & 91.67 & 1 & Household + body parts\\
2017 & SURF + k-mean + SVM \cite{Albini2017HumanInteractions} & Array with 768 tactels & 2 & 98.15 & 1 & Body parts \\
2017 & CNN \cite{Albini2017HumanInteractions} & Array with 768 tactels & 2 & 98.33 & 1 & Body parts \\
2017 & SVM \cite{lee2017learning} & NTU five-finger hand & 2 & 96.67 & 1 & Cylindrical, spherical or cubes \\
2018 & ROTConvPCE-mv \cite{Cao2018End-to-EndEnsemble} & Multiple hardware & 5 - 10 & 81.66 - 99.03 & Sequential data & Household \\
2018 & LDS-FCM \cite{LIU2018LDS-FCM:Recognition} & Multiple hardware & 5 - 10 & 91 - 100 & 1 & Household \\
\midrule
2018 & Transfer Learning (VGG16-SVM) & $[28 \times 50]$ Tactile array & 22 & $93.50$ & 1 & Household + body parts\\
2018 & Transfer Learning (ResNet-NN) & $[28 \times 50]$ Tactile array & 22 & $95.36$ & 1 & Household + body parts\\
2018 & TactNet-6 & $[28 \times 50]$ Tactile array & 22 & $95.02$ & 1 & Household + body parts \\
\bottomrule
\end{tabularx}
\end{table*}

\added{With the above considerations}, the CNN-based methods presented in this paper 
\added{offers advantages} with respect to the the state-of-the-art. In fact, this is the second research work with the largest number of objects in comparison with the state-of-the-art. Only \cite{Cretu2015ComputationalRecognition} has more classes, with 25 diverse symbols (letters), so that they may be easier to classify. Besides, the methods presented in this paper obtain the highest accuracy of all of those works that have more than 2 classes, with the exception of \cite{Cretu2015ComputationalRecognition}. Our proposals perform better than other methods with one order of magnitude less in the number of classes \added{and using only one touch}.

%
%
\section{Conclusion}
\label{sec:conclusions}

\added{
CNN-based methods for tactile object recognition based on high-resolution pressure images, to classify objects along with human-body parts (22 classes) using only one touch, have been presented, evaluated and compared. 
Tactile images have some similarities to video images, but have strong differences in depth-of-field, resolution, apart from other qualitative considerations that have been discusses in this work.
%
A custom-built CNN with a layer size that matches the size of the tactile image, and has been trained from scratch, in opposition to the use of transfer learning methods, that employ pre-trained CNNs in large video image dataset. In this approach, the classification layers of a CNN are replaced by an SVM or fine-tuned and re-trained in a set of up-sized tactile images with good recognition rates but a high computational cost.
Comparative experiments with 11 deep-leaning methods (8 transfer learning-based and 3 TactNet-based), have been carried out using the same tactile image dataset on the same hardware to get accuracy and evaluation times, showing the importance of using input layers with just the size of the sensor. Similar recognition rates with a huge reduction of classification time are be achieved by TactNet. Full training of from-scratch networks is a worthwhile one-time operation.
Other recent methods (such as SURF-SVM) applied to the recognition of tactile images have been qualitative and quantitatively compared with the presented approach without increasing the accuracy over the deep leaning methods.
A further reduction in the size of the tactile images for recognition with 
the presented approaches has been experimentally evaluated with different resolutions showing that for our dataset, a significant reduction in the number of tactels of the pressure image can be done while keeping a high recognition rate. This is important because a reduction in complexity and cost of the tactile sensor and computational power can be done for this kind of images.
Finally, tactile sensing is active, therefore, it requires physical interaction between the sensing surface and the environment. In robotics, the use of series of tactile information based on force/displacement will be considered. Considering sequences of tactile images, dynamic information of the contact can be used as input to the CNN to classify and identify physical properties of objects. In this case the performance and computational requirements of the 3D CNNs will be studied optimized to provide real-time execution.}



\ifCLASSOPTIONcaptionsoff
  \newpage
\fi


\par
\bibliographystyle{IEEEtran}
\bibliography{ms.bib}

\begin{thebibliography}{10}
\providecommand{\url}[1]{#1}
\csname url@samestyle\endcsname
\providecommand{\newblock}{\relax}
\providecommand{\bibinfo}[2]{#2}
\providecommand{\BIBentrySTDinterwordspacing}{\spaceskip=0pt\relax}
\providecommand{\BIBentryALTinterwordstretchfactor}{4}
\providecommand{\BIBentryALTinterwordspacing}{\spaceskip=\fontdimen2\font plus
\BIBentryALTinterwordstretchfactor\fontdimen3\font minus
  \fontdimen4\font\relax}
\providecommand{\BIBforeignlanguage}[2]{{%
\expandafter\ifx\csname l@#1\endcsname\relax
\typeout{** WARNING: IEEEtran.bst: No hyphenation pattern has been}%
\typeout{** loaded for the language `#1'. Using the pattern for}%
\typeout{** the default language instead.}%
\else
\language=\csname l@#1\endcsname
\fi
#2}}
\providecommand{\BIBdecl}{\relax}
\BIBdecl

\bibitem{LuoShanBimboJoaoDahiyaRavinder2017RoboticReview}
S.~Luo, J.~Bimbo, R.~Dahiya, and H.~Liu, ``Robotic tactile perception of object
  properties: A review,'' \emph{Mechatronics}, vol.~48, pp. 54--67, 2017.

\bibitem{Trujillo-Leon2018TactileWheelchairs}
A.~Trujillo-Leon, W.~Bachta, and F.~Vidal-Verdu, ``{Tactile Sensor-Based
  Steering as a Substitute of the Attendant Joystick in Powered Wheelchairs},''
  \emph{IEEE Transactions on Neural Systems and Rehabilitation Engineering},
  vol.~26, no.~7, pp. 1381--1390, 7 2018.

\bibitem{Bartolozzi2016RobotsTouch}
C.~Bartolozzi, L.~Natale, F.~Nori, and G.~Metta, ``{Robots with a sense of
  touch},'' \emph{Nature Materials}, vol.~15, no.~9, pp. 921--925, 9 2016.

\bibitem{Jamone2015HighlyHand}
L.~Jamone, L.~Natale, G.~Metta, and G.~Sandini, ``{Highly Sensitive Soft
  Tactile Sensors for an Anthropomorphic Robotic Hand},'' \emph{IEEE Sensors
  Journal}, vol.~15, no.~8, pp. 4226--4233, 8 2015.

\bibitem{Roncone2016PeripersonalSkin}
A.~Roncone, M.~Hoffmann, U.~Pattacini, L.~Fadiga, and G.~Metta, ``{Peripersonal
  space and margin of safety around the body: Learning visuo-tactile
  associations in a humanoid robot with artificial skin},'' \emph{PLoS ONE},
  vol.~11, no.~10, p. e0163713, 10 2016.

\bibitem{Dahiya2010TactileHumanoids}
R.~S. Dahiya, G.~Metta, M.~Valle, and G.~Sandini, ``{Tactile sensing-from
  humans to humanoids},'' \emph{IEEE Transactions on Robotics}, vol.~26, no.~1,
  pp. 1--20, 2 2010.

\bibitem{Gandarias2018EnhancingInteraction}
J.~M. Gandarias, J.~M. G{\'{o}}mez-de Gabriel, and A.~J. Garc{\'{i}}a-Cerezo,
  ``{Enhancing Perception with Tactile Object Recognition in Adaptive Grippers
  for Human–Robot Interaction},'' \emph{Sensors}, vol.~18, no.~3, p. 692, 2
  2018.

\bibitem{Chitta2011TactileManipulation}
S.~Chitta, J.~Sturm, M.~Piccoli, and W.~Burgard, ``{Tactile sensing for mobile
  manipulation},'' \emph{IEEE Transactions on Robotics}, vol.~27, no.~3, pp.
  558--568, 6 2011.

\bibitem{James2018SlipSensor}
J.~W. James, N.~Pestell, and N.~F. Lepora, ``{Slip Detection With a Biomimetic
  Tactile Sensor},'' \emph{IEEE Robotics and Automation Letters}, vol.~3,
  no.~4, pp. 3340--3346, 10 2018.

\bibitem{Romeo2017SlippageSensors}
R.~Romeo, C.~Oddo, M.~Carrozza, E.~Guglielmelli, and L.~Zollo, ``{Slippage
  Detection with Piezoresistive Tactile Sensors},'' \emph{Sensors}, vol.~17,
  no.~8, p. 1844, 8 2017.

\bibitem{Gandarias2017HumanSensor}
J.~M. Gandarias, J.~M. Gomez-de Gabriel, and A.~Garcia-Cerezo, ``{Human and
  object recognition with a high-resolution tactile sensor},'' in \emph{IEEE
  Sensors Conference}, 2017.

\bibitem{Luo2016IterativeRecognition}
S.~Luo, W.~Mou, K.~Althoefer, and H.~Liu, ``{Iterative Closest Labeled Point
  for Tactile Object Shape Recognition},'' in \emph{IEEE/RSJ International
  Conference on Intelligent Robots and Systems (IROS)}, 2016.

\bibitem{Yuan2017DesignClassification}
Q.~Yuan and J.~Wang, ``{Design and Experiment of the NAO Humanoid Robot's
  Plantar Tactile Sensor for Surface Classification},'' in \emph{4th
  International Conference on Information Science and Control Engineering
  (ICISCE)}, 2017.

\bibitem{Hoelscher2015EvaluationRecognition}
J.~Hoelscher, J.~Peters, and T.~Hermans, ``{Evaluation of tactile feature
  extraction for interactive object recognition},'' in \emph{IEEE-RAS
  International Conference on Humanoid Robots}, 2015.

\bibitem{li2013sensing}
R.~Li and E.~H. Adelson, ``Sensing and recognizing surface textures using a
  gelsight sensor,'' in \emph{IEEE Conference on Computer Vision and Pattern
  Recognition (CVPR)}, 2013, pp. 1241--1247.

\bibitem{vidal2011three}
F.~Vidal-Verd{\'u}, {\'O}.~Oballe-Peinado, J.~A. S{\'a}nchez-Dur{\'a}n,
  J.~Castellanos-Ramos, and R.~Navas-Gonz{\'a}lez, ``Three realizations and
  comparison of hardware for piezoresistive tactile sensors,'' \emph{Sensors},
  vol.~11, no.~3, pp. 3249--3266, 2011.

\bibitem{Chathuranga2016MagneticSensor}
D.~S. Chathuranga, Z.~Wang, Y.~Noh, T.~Nanayakkara, and S.~Hirai, ``{Magnetic
  and Mechanical Modeling of a Soft Three-Axis Force Sensor},'' \emph{IEEE
  Sensors Journal}, vol.~16, no.~13, pp. 5298--5307, 7 2016.

\bibitem{Ward-Cherrier2018TheMorphologies}
B.~Ward-Cherrier, N.~Pestell, L.~Cramphorn, B.~Winstone, M.~E. Giannaccini,
  J.~Rossiter, and N.~F. Lepora, ``{The TacTip Family: Soft Optical Tactile
  Sensors with 3D-Printed Biomimetic Morphologies},'' \emph{Soft Robotics},
  vol.~5, no.~2, pp. 216--227, 4 2018.

\bibitem{Gong2017ARobots}
D.~Gong, R.~He, J.~Yu, and G.~Zuo, ``{A pneumatic tactile sensor for
  co-operative robots},'' \emph{Sensors}, vol.~17, no.~11, p. 2592, 11 2017.

\bibitem{Maiolino2013ARobots}
P.~Maiolino, M.~Maggiali, G.~Cannata, G.~Metta, and L.~Natale, ``{A Flexible
  and Robust Large Scale Capacitive Tactile System for Robots},'' \emph{IEEE
  Sensors Journal}, vol.~13, no.~10, pp. 3910--3917, 10 2013.

\bibitem{YiZhengkun}
Z.~Yi, R.~Calandra, F.~Veiga, H.~van Hoof, T.~Hermans, Y.~Zhang, and J.~Peters,
  ``Active tactile object exploration with gaussian processes,'' in
  \emph{IEEE/RSJ International Conference on Intelligent Robots and Systems
  (IROS)}, 2016.

\bibitem{Corradi2015BayesianSensor}
T.~Corradi, P.~Hall, and P.~Iravani, ``{Bayesian tactile object recognition:
  Learning and recognising objects using a new inexpensive tactile sensor},''
  in \emph{IEEE International Conference on Robotics and Automation (ICRA)},
  2015.

\bibitem{Albini2017HumanInteractions}
A.~Albini, S.~Denei, and G.~Cannata, ``{Human Hand Recognition From Robotic
  Skin Measurements in Human-Robot Physical Interactions},'' in \emph{IEEE/RSJ
  International Conference on Intelligent Robots and Systems (IROS)}, 2017.

\bibitem{Luo2015NovelRecognition}
S.~Luo, W.~Mou, K.~Althoefer, and H.~Liu, ``{Novel Tactile-SIFT Descriptor for
  Object Shape Recognition},'' \emph{IEEE Sensors Journal}, vol.~15, no.~9, pp.
  5001--5009, 9 2015.

\bibitem{Krizhevsky2012ImageNetNetworks}
A.~Krizhevsky, I.~Sutskever, and G.~E. Hinton, ``{ImageNet Classification with
  Deep Convolutional Neural Networks},'' \emph{Advances In Neural Information
  Processing Systems}, pp. 1--9, 2012.

\bibitem{Cao2018End-to-EndEnsemble}
L.~Cao, F.~Sun, X.~Liu, W.~Huang, R.~Kotagiri, and H.~Li, ``{End-to-End ConvNet
  for Tactile Recognition Using Residual Orthogonal Tiling and Pyramid
  Convolution Ensemble},'' \emph{Cognitive Computation}, pp. 1--19, 6 2018.

\bibitem{Shibata2017}
A.~Shibata, A.~Ikegami, M.~Nakauma, and M.~Higashimori, ``{Convolutional Neural
  Network based Estimation of Gel-like Food Texture by a Robotic Sensing
  System},'' \emph{Robotics}, vol.~6, no.~4, p.~37, 12 2017.

\bibitem{gandarias2017tactile}
J.~M. Gandarias, J.~M. G{\'o}mez-de Gabriel, and A.~J. Garc{\'\i}a-Cerezo,
  ``Tactile sensing and machine learning for human and object recognition in
  disaster scenarios,'' in \emph{Third Iberian Robotics conference}.\hskip 1em
  plus 0.5em minus 0.4em\relax Springer, 2017.

\bibitem{pan2010survey}
S.~J. Pan, Q.~Yang \emph{et~al.}, ``A survey on transfer learning,'' \emph{IEEE
  Transactions on knowledge and data engineering}, vol.~22, no.~10, pp.
  1345--1359, 2010.

\bibitem{mihalkova2009transfer}
L.~Mihalkova and R.~J. Mooney, ``Transfer learning from minimal target data by
  mapping across relational domains.'' in \emph{International Joint Conference
  on Artificial Intelligence (IJCAI)}, 2009.

\bibitem{Feng2018ActiveObjects}
D.~Feng, M.~Kaboli, and G.~Cheng, ``{Active Prior Tactile Knowledge Transfer
  for Learning Tactual Properties of New Objects},'' \emph{Sensors}, vol.~18,
  no.~2, p. 634, 2 2018.

\bibitem{Kaboli2018RobustSkin}
M.~Kaboli and G.~Cheng, ``{Robust Tactile Descriptors for Discriminating
  Objects From Textural Properties via Artificial Robotic Skin},'' \emph{IEEE
  Transactions on Robotics}, pp. 1--19, 2018.

\bibitem{Baishya2016RobustLearning}
S.~S. Baishya and B.~Bauml, ``{Robust material classification with a tactile
  skin using deep learning},'' in \emph{IEEE/RSJ International Conference on
  Intelligent Robots and Systems (IROS)}, 2016.

\bibitem{Madry2014ST-HMP:Data}
M.~Madry, L.~Bo, D.~Kragic, and D.~Fox, ``{ST-HMP: Unsupervised Spatio-Temporal
  feature learning for tactile data},'' in \emph{IEEE International Conference
  on Robotics and Automation (ICRA)}, 2014.

\bibitem{Liu2016ObjectMethods}
H.~Liu, D.~Guo, and F.~Sun, ``{Object Recognition Using Tactile Measurements:
  Kernel Sparse Coding Methods},'' \emph{IEEE Transactions on Instrumentation
  and Measurement}, vol.~65, no.~3, pp. 656--665, 3 2016.

\bibitem{kerzel}
M.~Kerzel, M.~Ali, H.~G. Ng, and S.~Wermter, ``{Haptic material classification
  with a multi-channel neural network},'' in \emph{International Joint
  Conference on Neural Networks (IJCNN)}, 2017.

\bibitem{jamali2011majority}
N.~Jamali and C.~Sammut, ``Majority voting: Material classification by tactile
  sensing using surface texture,'' \emph{IEEE Transactions on Robotics},
  vol.~27, no.~3, pp. 508--521, 2011.

\bibitem{Liu2012ASensor}
H.~Liu, X.~Song, T.~Nanayakkara, L.~D. Seneviratne, and K.~Althoefer, ``{A
  computationally fast algorithm for local contact shape and pose
  classification using a tactile array sensor},'' in \emph{IEEE International
  Conference on Robotics and Automation (ICRA)}, 2012.

\bibitem{Martinez-Hernandez2017FeelingHand}
U.~Martinez-Hernandez, T.~J. Dodd, and T.~J. Prescott, ``{Feeling the Shape:
  Active Exploration Behaviors for Object Recognition With a Robotic Hand},''
  \emph{IEEE Transactions on Systems, Man, and Cybernetics: Systems}, pp.
  1--10, 2017.

\bibitem{Khasnobish2012Object-shapeNetwork}
A.~Khasnobish, A.~Jati, G.~Singh, S.~Bhattacharyya, A.~Konar, D.~Tibarewala,
  E.~Kim, and A.~K. Nagar, ``{Object-shape recognition from tactile images
  using a feed-forward neural network},'' in \emph{International Joint
  Conference on Neural Networks (IJCNN)}, 2012.

\bibitem{Schmitz2014TactileDropout}
A.~Schmitz, Y.~Bansho, K.~Noda, H.~Iwata, T.~Ogata, and S.~Sugano, ``{Tactile
  object recognition using deep learning and dropout},'' in \emph{IEEE-RAS
  International Conference on Humanoid Robots}, 2014.

\bibitem{Lawrence1997FaceApproach}
S.~Lawrence, C.~L. Giles, A.~C. Tsoi, and A.~D. Back, ``{Face recognition: A
  convolutional neural-network approach},'' \emph{IEEE Transactions on Neural
  Networks}, vol.~8, no.~1, pp. 98--113, 1997.

\bibitem{Falco2017Cross-modalExploration}
P.~Falco, S.~Lu, A.~Cirillo, C.~Natale, S.~Pirozzi, and D.~Lee, ``{Cross-modal
  visuo-tactile object recognition using robotic active exploration},'' in
  \emph{IEEE International Conference on Robotics and Automation (ICRA)}, 2017.

\bibitem{gao2016deep}
Y.~Gao, L.~A. Hendricks, K.~J. Kuchenbecker, and T.~Darrell, ``Deep learning
  for tactile understanding from visual and haptic data,'' in \emph{IEEE
  International Conference on Robotics and Automation (ICRA)}, 2016.

\bibitem{Zheng2016DeepInformation}
H.~Zheng, L.~Fang, M.~Ji, M.~Strese, Y.~Ozer, and E.~Steinbach, ``{Deep
  Learning for Surface Material Classification Using Haptic and Visual
  Information},'' \emph{IEEE Transactions on Multimedia}, vol.~18, no.~12, pp.
  2407--2416, 12 2016.

\bibitem{luo2015tactile}
S.~Luo, X.~Liu, K.~Althoefer, and H.~Liu, ``Tactile object recognition with
  semi-supervised learning,'' in \emph{International Conference on Intelligent
  Robotics and Applications}, 2015.

\bibitem{schneider2009object}
A.~Schneider, J.~Sturm, C.~Stachniss, M.~Reisert, H.~Burkhardt, and W.~Burgard,
  ``Object identification with tactile sensors using bag-of-features,'' in
  \emph{IEEE/RSJ International Conference on Intelligent Robots and Systems
  (IROS)}, 2009.

\bibitem{Luo2014RotationSensor}
S.~Luo, W.~Mou, M.~Li, K.~Althoefer, and H.~Liu, ``{Rotation and Translation
  Invariant Object Recognition with a Tactile Sensor},'' in \emph{IEEE Sensors
  Conference}, 2014.

\bibitem{Hisham2017AnMethod}
M.~Hisham, S.~N. Yaakob, R.~Raof, A.~Nazren, and N.~Wafi, ``An analysis of
  performance for commonly used interpolation method,'' \emph{Advanced Science
  Letters}, vol.~23, no.~6, pp. 5147--5150, 2017.

\bibitem{he2016deep}
K.~He, X.~Zhang, S.~Ren, and J.~Sun, ``Deep residual learning for image
  recognition,'' in \emph{Proceedings of the IEEE Conference on Computer Vision
  and Pattern Recognition (CVPR)}, 2016.

\bibitem{iandola2016squeezenet}
F.~N. Iandola, S.~Han, M.~W. Moskewicz, K.~Ashraf, W.~J. Dally, and K.~Keutzer,
  ``Squeezenet: Alexnet-level accuracy with 50x fewer parameters and< 0.5 mb
  model size,'' \emph{arXiv preprint arXiv:1602.07360}, 2016.

\bibitem{Simonyan2014VeryRecognition}
K.~Simonyan and A.~Zisserman, ``Very deep convolutional networks for
  large-scale image recognition,'' \emph{arXiv preprint arXiv:1409.1556}, 2014.

\bibitem{russakovsky2015imagenet}
O.~Russakovsky, J.~Deng, H.~Su, J.~Krause, S.~Satheesh, S.~Ma, Z.~Huang,
  A.~Karpathy, A.~Khosla, M.~Bernstein \emph{et~al.}, ``Imagenet large scale
  visual recognition challenge,'' \emph{International Journal of Computer
  Vision}, vol. 115, no.~3, pp. 211--252, 2015.

\bibitem{kauderer2017quantifying}
E.~Kauderer-Abrams, ``Quantifying translation-invariance in convolutional
  neural networks,'' \emph{arXiv preprint arXiv:1801.01450}, 2017.

\bibitem{Canziani2016AnApplications}
A.~Canziani, A.~Paszke, and E.~Culurciello, ``An analysis of deep neural
  network models for practical applications,'' \emph{arXiv preprint
  arXiv:1605.07678}, 2016.

\bibitem{Navarro2012HapticHands}
S.~E. Navarro, N.~Gorges, H.~W{\"o}rn, J.~Schill, T.~Asfour, and R.~Dillmann,
  ``Haptic object recognition for multi-fingered robot hands,'' in
  \emph{Haptics Symposium}, 2012.

\bibitem{Cretu2015ComputationalRecognition}
A.~M. Cretu, T.~E.~A. De~Oliveira, V.~Prado Da~Fonseca, B.~Tawbe, E.~M. Petriu,
  and V.~Z. Groza, ``{Computational intelligence and mechatronics solutions for
  robotic tactile object recognition},'' in \emph{WISP 2015 - IEEE
  International Symposium on Intelligent Signal Processing, Proceedings}, 2015.

\bibitem{lee2017learning}
W.~Y. Lee, M.~B. Huang, and H.~P. Huang, ``Learning robot tactile sensing of
  object for shape recognition using multi-fingered robot hands,'' in
  \emph{26th IEEE International Symposium on Robot and Human Interactive
  Communication (RO-MAN)}, 2017.

\bibitem{LIU2018LDS-FCM:Recognition}
C.~Liu, W.~Huang, F.~Sun, M.~Luo, and C.~Tan, ``{LDS-FCM: A Linear Dynamical
  System-based Fuzzy C-Means Method for Tactile Recognition},'' \emph{IEEE
  Transactions on Fuzzy Systems}, 2018.

\end{thebibliography}

\vfill

\end{document}